\documentclass[10pt,twocolumn,letterpaper]{article}

\usepackage{iccv}
\usepackage{times}
\usepackage{epsfig}
\usepackage{graphicx}
\usepackage{amsmath}
\usepackage{amssymb}
\usepackage{booktabs} 
\usepackage{multirow}
\usepackage{pifont}
\usepackage{makecell}
\usepackage{lipsum}

\usepackage[pagebackref=true,breaklinks=true,letterpaper=true,colorlinks,bookmarks=false]{hyperref}

\iccvfinalcopy

\begin{document}

\title{On the Significance of Question Encoder Sequence Model \\ in the Out-of-Distribution Performance in Visual Question Answering}

\author{Gouthaman KV \ \ \ \ Anurag Mittal\\\\
Indian Institute of Technology Madras, India\\
{\tt\small \{gkv,amittal\}@cse.iitm.ac.in}
}

\maketitle

\newcommand\blfootnote[1]{%
  \begingroup
  \renewcommand\thefootnote{}\footnote{#1}%
  \addtocounter{footnote}{-1}%
  \endgroup
}

\begin{abstract}
   Generalizing beyond the experiences has a significant role in developing practical AI systems. It has been shown that current Visual Question Answering (VQA) models are over-dependent on the language-priors (spurious correlations between question-types and their most frequent answers) from the train set and pose poor performance on Out-of-Distribution (OOD) test sets. This conduct limits their generalizability and restricts them from being utilized in real-world situations.  This paper shows that the sequence model architecture used in the question-encoder has a significant role in the generalizability of VQA models. To demonstrate this, we performed a detailed analysis of various existing  RNN-based and Transformer-based question-encoders, and along, we proposed a novel Graph attention network (GAT)-based question-encoder.  Our study found that a better choice of sequence model in the question-encoder improves the generalizability of VQA models even without using any additional relatively complex bias-mitigation approaches.
\end{abstract}


\section{Introduction}
Visual Question Answering (VQA), i.e., answering questions about images, is a strong benchmark task for context-specific reasoning and image understanding. Solving VQA involves the interaction between the vision and language modalities and allows various challenging real-world applications, such as aiding visually-impaired users to understand their surroundings, interact with robots in natural language, question-based image retrieval, etc.  Since the VQA task measures the machine's ability to understand the images via a natural language query, this multi-modal task garnered wide attention from both the computer vision and natural language processing (NLP) communities. As a result, several VQA models have been proposed to solve the task~\cite{qiwu_survey,kafle_cviu,lxmert,lat}.
A VQA model is intended to predict the answer based on the visual evidences from the image and the question's actual intention. Unfortunately, often this is not the case even with state-of-the-art (SOTA) VQA models. Instead of predicting the right answers for the right reasons, they often learn shortcuts by exploiting the ``language-priors" (superficial statistical correlations between the question-types to their most frequent answers) in the train set~\cite{gvqa,overcoming,rubi}.  For instance, most of the models overwhelmingly answer the questions ``How many X" with ``2",  ``Is the X," with ``Yes," ``What color is X" with ``White," etc., irrespective of the question part ``X."  
This over-dependency of the models on the language-priors causes poor performance on Out-of-Distribution (OOD) test sets that pose different language-priors than the train set~\cite{gvqa}. As such, evaluation on OOD test sets has emerged as a proxy for measuring generalization.   Thus the negative effect of language-priors will become potentially detrimental to creating a general VQA system, as generalizing beyond the experiences has a significant role in developing practical AI systems.

Recently, various bias-mitigation approaches have been proposed to reduce the effect of language-priors and improve the OOD performance in VQA.  
In particular, some approaches focus on reducing the question over-fitting with various regularization techniques using additional trainable question-only branches~\cite{overcoming,rubi}. Some methods try to improve the model's visual-grounding by training using additional manually annotated data~\cite{hint,self_critical_bias}. At the same time, some other approaches use unbiased training methods that use various data augmentation techniques. Such approaches manipulate the training data and generate additional counter-factual training samples~\cite{css,mutant}. It is shown that using the counter-factual samples in addition to the original training data helps to reduce the effect of language-priors and improve the OOD performance. However, a common downside of all the above approaches is that they require additional complex engineering on top of existing models and cause notable performance degradation on in-distribution test sets while improving OOD performance.

The widely-adopted architecture pipeline for VQA is the encoder-decoder approach, where first, the modalities are encoded separately, and the higher-order interactions between the encoded features are modeled later~\cite{kafle_cviu,qiwu_survey}. Since the question is a sequence of words, a visible trend in the VQA literature is that most of the existing models pre-determine the question-encoder as the sequence model architecture as used in the SOTA language models in the NLP community at that time. Due to this, RNN-based question encoders were the prominent architecture choice for quite a long time~\cite{qiwu_survey,bottomup,murel,ban}, and more recently, proposed models started following the Transformer-based encoders~\cite{vilbert,lxmert,vlbert}.   As a result, the question encoder is seldom studied in the VQA literature, as most models focus on the image side or interactions between the encoded question and the image.  Existing bias-mitigation approaches in VQA also overlook the question-encoder sequence model and focus on various other aspects to mitigate the bias.  

In contrast to this, in this paper, we show that the sequence model in the question-encoder itself has a significant role in the OOD performance in VQA.  We found that a better sequence model in the question-encoder can push the VQA model to improve the OOD performance without any additional relatively complex bias-mitigation approaches.  To demonstrate this, we did a detailed analysis of various question-encoders based on different sequence model architectures. Our study includes analysis of various existing RNN-based and Transformer-based question-encoders and the proposal of a novel Graph attention network (GAT)-based question-encoder.   To the best of our knowledge, such a study is not yet reported in VQA, and ours mark the first of its kind.  We summarize the major findings from this study below: 

\begin{itemize}
     \item The sequence model in the question-encoder has a significant role in the OOD performance and generalizability of VQA models. A better sequence model choice can reduce over-fitting to language-priors and push the models to boost the generalizability without using additional complex bias-mitigation approaches.

     \item Self-attention-based question-encoders show better generalizability in VQA. Interestingly, among self-attention-based architectures, a single-layered GAT-based question-encoder shows better generalization than its multi-layered Transformer-based rivals.
 
     \item The currently popular learnable absolute ``position-encodings" to encode the word-order information in the Transformer-based question-encoders has a considerable role in over-fitting the VQA model to the language-priors.  The model primarily uses such information to learn spurious easy routes exploiting the language-priors and benefit on the in-distribution performance.  We found that a ``1-D CNN" is a better alternative to the ``position-encodings".

\end{itemize}

Based on our findings, we believe that, in VQA, the sequence model in the question-encoder needs more attention and care instead of considering it as a black-box using the sequence model as used in the  SOTA language modeling settings from the NLP community.   We hope this study can significantly advance the VQA research in developing robust models resilient to language-priors as prior bias-mitigation approaches are indifferent to the question-encoder sequence model~\cite{overcoming,self_critical_bias,rubi,bias_aaai,hint,css,mutant}.

\section{Related works}
\textbf{Existing bias-mitigation approaches in VQA:}
A direct solution to reduce the language-priors is to balance the train set with more visual data.
In this regard, \cite{vqa2}~included complementary images (that look similar to the questioned image but with a different answer to the same question) to the VQAv1 dataset~\cite{vqa1} and proposed the more balanced VQAv2 dataset. \cite{gvqa}~demonstrated that the VQAv2 train\&test sets have similar language-priors. Owing to this, leveraging such priors from the train set will still benefit the model while testing. To this end, they proposed a new dataset, the VQA-CP (VQA under Changing Priors), which is designed to test the models in an OOD setting by providing significantly different language-priors between the train and test sets. Currently, this dataset is widely used to measure a VQA model's capacity to be resilient to language-priors and improve generalizability.

Apart from the dataset side, various methods have been developed to overcome the language-priors. \cite{gvqa} proposed a specific VQA model (GVQA) with various architectural regulations to reduce the effect of language-priors. \cite{overcoming,rubi} proposed various regularization schemes by adding an auxiliary question-only branch to the model. This branch is intended to capture the language-priors. Then, ~\cite{overcoming} used gradient negation on the question-only branch and trained the whole model in an adversarial manner. \cite{rubi} masked the output from the question-only branch with the ``sigmoid" operation. Then the masked output is fused with the original model output to adapt the loss to compensate for the language-priors dynamically. \cite{hint} proposed a tuning approach, using additional manually-annotated attention maps from the VQA-HAT dataset~\cite{vqa-hat}. This additional tuning helps the model to improve the visual-grounding capacity and thus reduces the over-dependency on the language-priors. Similar to this approach, \cite{self_critical_bias} proposed another tuning approach that makes the model decide if the focused image region to answer the question is correct or not. For this, the model is provided with additional manually-annotated attention-maps from VQA-HAT dataset~\cite{vqa-hat} or textual explanations from VQA-X dataset~\cite{vqa-x}. Another notable prior work is
\cite{ensemble_lmh}, where the authors proposed an ensemble-based training scheme. Their method has two steps: in the first step, a bias-only model is constructed by taking only the question-type information (e.g., ``What color is" 
) as the input. Then this bias-only model prediction is used as a training constraint for the second step, where the best model is chosen from an ensemble of models to predict the final answer. \cite{css,mutant} proposed various data augmentation methods to generate additional counterfactual training samples to train the model without exploiting the language-priors. \cite{vgqe}, \cite{bias_aaai} tries to improve the encoded question representation to reduce the language-priors.  \cite{vgqe} try to improve the visual-grounding of the model by generating visually-grounded question representations. They found that fusing visual information to the question-words reduces the over-dependency of the model on the language side. \cite{bias_aaai} split the question into three language phrases and use an attention mechanism to regularize the contribution from each of the phrases.

All of the approaches mentioned above do not pay attention to the sequence model in the question-encoder.  Instead, all of them follow the same sequence model used in the question-encoder in the respective baselines that they experimented with, then use additional techniques to overcome the language-priors. In contrast, we show that the sequence model choice in the question-encoder has a significant role in the generalizability of VQA models, and a better sequence model improves the OOD performance and generalizability of VQA models.

\textbf{Behaviour analysis of VQA models:}
Our work is related to~\cite{analyzing_vqa} and~\cite{did_model} in terms of the nature of the study.  \cite{analyzing_vqa} analyzed VQA models that use the RNN-based question-encoders and found that such models mostly focus on the beginning of the question and often overlook the rest of the parts. Similar to this, \cite{did_model} did several tests on VQA models that use the RNN-based question encoders and found that they often ignore important question words. Our work lies on the same line as these works.  In contrast to the above works, we studied various sequence model architectures in the question-encoder and analyzed their impact on the OOD performance of VQA models.  Our study found that the sequence model in the question-encoder significantly affects the OOD performance in VQA.  To the best of our knowledge, such a study is not yet reported in the VQA literature.  We hope this study will inspire the community to focus more on the question-encoder to develop robust VQA models that are resilient to language-priors.

\textbf{Graph Neural Networks as question-encoder:}
In VQA, most of the existing models use RNN-based~\cite{bottomup,murel,ban} or Transformer-based~\cite{lxmert,vilbert,mlin,mca} question-encoders. However, we are aware of few models that use the question syntax-graph obtained from off-the-shelf pre-trained language parsers and use Graph neural networks (GNN) to encode the question. For instance, \cite{wheretolook} categorized the words into several bins based on the parser output, and word-embeddings from each bin are concatenated to get the encoded question. \cite{Teney_graph} use first-order Gated Graph Sequence Neural Network (GGS-NN)~\cite{gated_graph} on top of the parser output to encode the question. One major limitation of these approaches is that the question pre-processing with an external pre-trained language-parser is a time-consuming process and increases test time. In contrast to these approaches, we propose a novel question-encoder GAT-QE that considers the question as a complete digraph and uses a Graph attention network (GAT) and a 1-D CNN to encode the question. The GAT-QE does not require the use of any off-the-shelf language parsers. As per our understanding, this is the first paper to explore the use of ``GAT+1-D CNN" as a question-encoder.

\section{Analyzed question encoders} \label{sec_qe}
This section explains the various types of question encoders that we have considered in this paper.  It includes the question-encoders from various existing best performing VQA models (i.e., RNN-based and Transformer-based) and the proposal of a novel Graph attention network-based question-encoder. Before going into the details of each question-encoder, we first explain the common notations used in all of them. The question is considered as a sequence of words. The set of word-level features is denoted as $Q=\{q_1,..,q_n\}$, where $q_i\in \mathcal{R}^{d_w}$ is the $i^{th}$ word feature and $n$ is the length of the sequence. The output of the question-encoder (i.e., the encoded question context vector) is denoted as $q\in \mathcal{R}^{d_q}$.

  \begin{figure}[t]
    \centering
    \includegraphics[width=0.8\columnwidth]{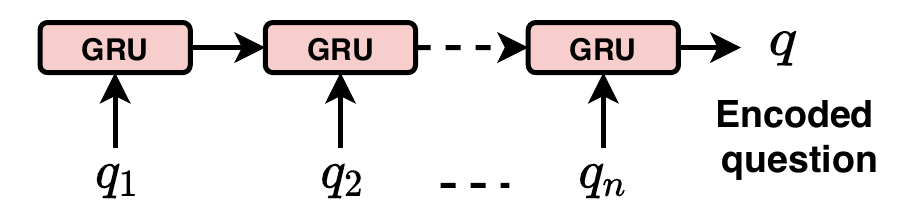}
    \caption{GRU-QE: The words are processed as in the same order in the question. $q_i$ is the word-embedding of the $i^{th}$ word. }
    \label{fig:rnnqe}
\end{figure}
\subsection{RNN-based}
Gated recurrent neural networks such as LSTMs and GRUs, were the prominent sequence model architecture for quite a long time in NLP~\cite{lstm_translation}. Since the question is a sequence of words, the gated RNNs naturally became the choice of question encoder in many existing VQA models~\cite{qiwu_survey,kafle_cviu}.
In this paper, we use the GRU to implement this type of question encoder, and we call it GRU-QE. An illustration of the GRU-QE is shown in Figure~\ref{fig:rnnqe}.  The GRU-QE processes the words in the same order as in the question. First, the words are converted into the set of word-level features $Q$ using some word-embedding networks (we use the Glove~\cite{glove} word-embeddings).  These features are passed through a sequence of GRU cells one-by-one in the same order as in the question.  Formally, we denote a GRU cell at time $t$ as $h_t=\text{GRU}(q_t,h_{t-1};\theta)$, 
where $q_t$ is the current question word embedding, $h_{t-1}$ and $h_t$  are the previous and current hidden states respectively, and $\theta$ is the learnable parameters of the GRU cell. Then, the final cell hidden state vector is considered as the encoded question, i.e., $q=h_n$.

   \begin{figure}[t]
    \centering
    \includegraphics[width=0.8\columnwidth]{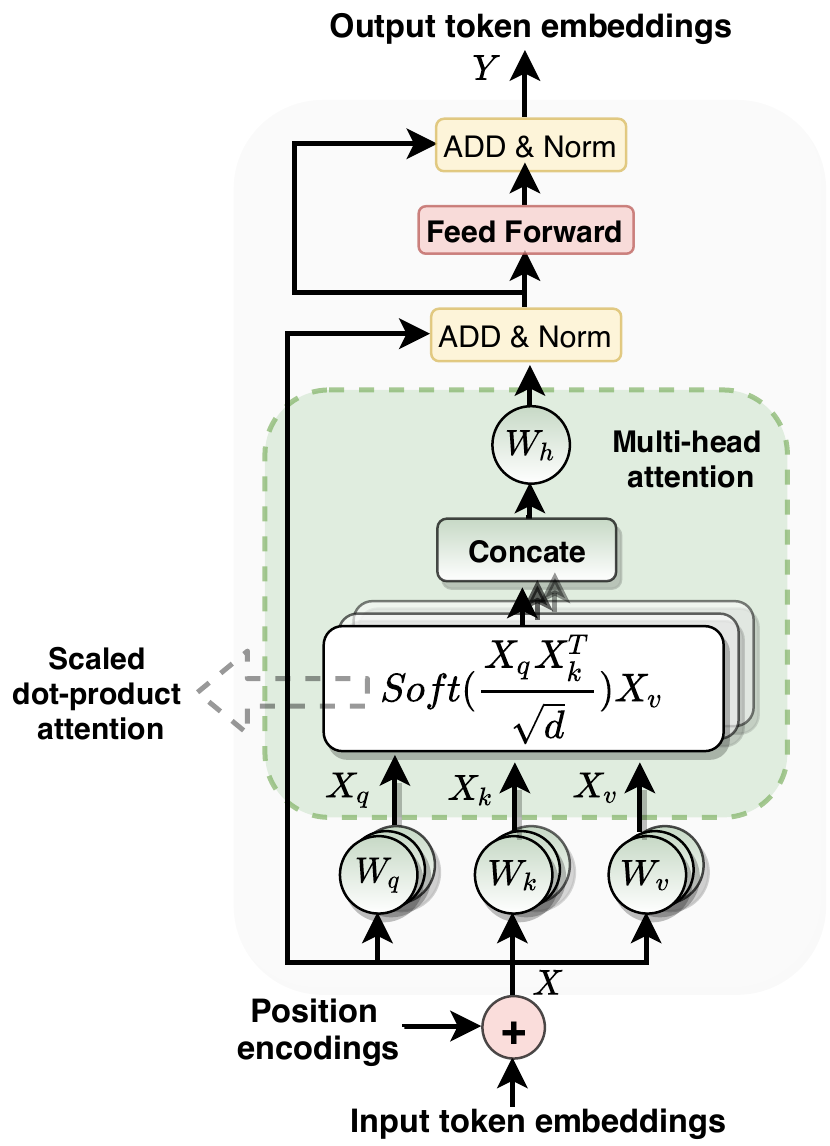}
    \caption{Illustration of a Transformer module. ``Soft" denotes Softmax, 
    $W_{(\cdot)}$ denotes linear transformations.}
    \label{fig:transformer}
\end{figure} 

\subsection{Transformer-based}
Currently, Transformer-based~\cite{transformer} architectures are the SOTA sequence models in various NLP tasks~\cite{bert}. An illustration of a Transformer module is shown in Figure~\ref{fig:transformer}. The Transformer considers the sequence as a collection of ``tokens" (individual elements of the sequence) and processes them parallelly, unlike in RNNs. To bring each token's absolute position information in the sequence, the Transformer uses special vectors called ``position-encodings". These vectors are added to the input token embeddings (feature vectors of the tokens) and pass through the Transformer module. It will output the updated representations for each input token. 
Refer to~\cite{transformer} for a complete mathematical description of each of the components in the Transformer module.  

In this paper, we use two types of Transformer-based question encoders (i.e., BERT-QE: Language-only pre-trained and MMT-QE: Vision-Language pre-trained) commonly used in VQA, described below. 

\begin{figure}[t]
    \centering
    \includegraphics[width=\linewidth]{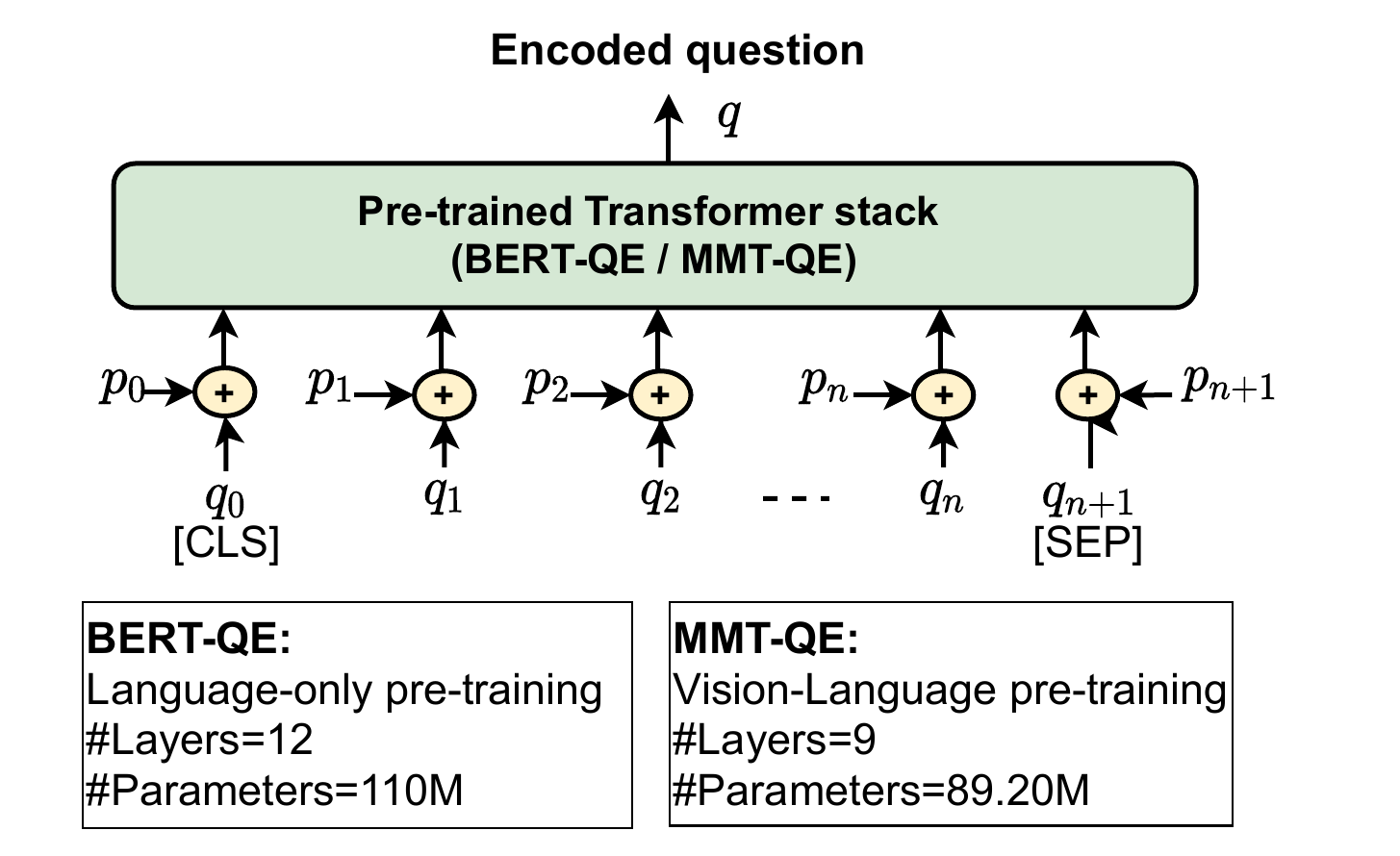}
    \caption{Transformer-based question encoders. The symbol $+$ indicates addition operation, $q_i$ and $p_i$ represents the $i^{th}$ input token embedding and position encoding vectors respectively. }
    \label{fig:transformer_qe}
\end{figure}

\textbf{BERT-QE:} BERT~\cite{bert} is a Language-only pre-trained Transformer-based sequence model, that showed SOTA performance in various language understanding tasks in NLP.  The main idea is to first pre-train a stack of Transformer modules on large Language-only data and then fine-tune them for several downstream language understanding tasks.  Inspired by the success of BERT in NLP, using it as the question-encoder became popular in VQA~\cite{mlin}.  The main idea is to use a Language-only pre-trained BERT model as the question-encoder and fine-tune it while training the VQA model.  In this paper, we use the pre-trained ``BERT-base" model as the question-encoder (we call it BERT-QE) for various experiments. The BERT-QE contains 12 Transformers and a total of 110M parameters, pre-trained on BooksCorpus~\cite{bookcorpus} and English Wikipedia.

    \textbf{MMT-QE: } 
    Another emerging trend in VQA is the use of Vision-Language pre-trained Transformers as the question encoders~\cite{vilbert,lxmert,vlbert}. The main idea is to pre-train a stack of Transformer modules on a large amount of Vision-Language data. Then, fine-tune them while training the VQA model. 
    This category of question-encoders are used in the current SOTA (in terms of in-distribution performance) VQA models~\cite{vilbert,lxmert}. We call this category of question-encoders as multi-modal Transformer-based question-encoders (MMT-QEs) as they are Transformers pre-trained using data from multiple modalities (i.e., vision \& language).  
    In this paper, we choose the question-encoder from the LXMERT model by~\cite{lxmert} as the representative of MMT-QE, for our experiments. It is a 9-layered MMT-QE containing a total of 89.20M parameters, pre-trained on five Vision-Language datasets consisting of 9.18M image-sentence pairs.

\begin{figure*}
    \centering
    \includegraphics[width=0.95\textwidth]{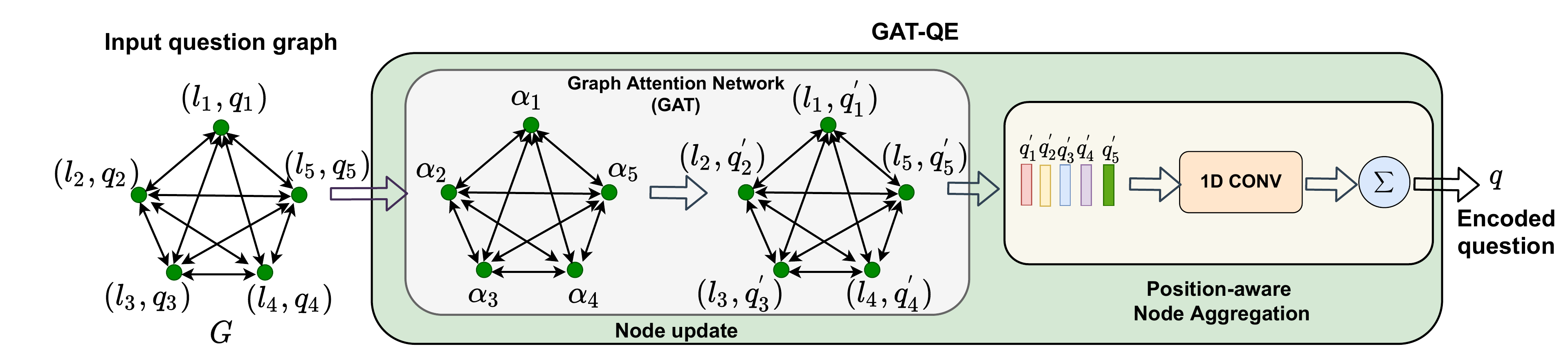}
    \caption{Illustration of GAT-QE on a question with five words. Self-loops are implicit in all the nodes of the graph $G$.}
    \label{fig:gatqe}
\end{figure*}

    A visual comparison of BERT-QE and MMT-QE is shown in Figure~\ref{fig:transformer_qe}. Both of them follow a similar working pipeline, as described below.  First, they appends two special tokens [CLS] and [SEP] respectively to the beginning and end of the sequence of question words.  Then, they use the WordPiece embeddings~\cite{wordpiece} to find the feature vectors of each token in the sequence.  This process results in the set of input token embeddings  $Q=\{q_0,q_1,..,q_{n},q_{n+1}\}$ where $q_0$ and $q_{n+1}$ correspond to the [CLS] and [SEP] tokens respectively, and all the others (i.e., $q_1,..,q_n$) correspond to the words from the question. Then, the ``position-encodings" ($p_i$) are added to each of the token embeddings ($q_i$) to bring the absolute position information (i.e., word-order information). Both BERT-QE and MMT-QE use the learned ``position-encodings", i.e., a set of learnable vectors, to indicate the absolute position.  The updated token embeddings (i.e., $q_i+p_i$) are then input to the  BERT-QE or MMT-QE. Then, all the final layer (i.e., $12^{th}$ in BERT-QE and $9^{th}$ in MMT-QE) Transformer's output token embeddings are sum-pooled to get the encoded question $q$.

\subsection{Graph attention network (GAT)-based}
Graph Convolutional Networks (GCNs)~\cite{gcn} were shown effective in various NLP tasks such as Semantic-role labeling~\cite{semantic_role}, Language translation~\cite{gcn_translation}, etc.  In principle, GCNs are flexible enough to capture all the linguistic structures, as long as they are represented as graphs~\cite{semantic_role,gcn_translation}. Inspired by this, we propose a novel question-encoder for VQA, the GAT-QE, that uses a Graph attention network (GAT)~\cite{gat} and a 1-D CNN to encode the question.  
GAT is a type of GCN where the node features are updated based on self-attention over the neighborhood nodes. 
The GAT-QE considers the question as a complete digraph (i.e., fully-connected directed graph) with words as nodes and edges denotes various syntactic connections.  This complete-digraph structure of the question provides the following advantages: 
1) Provides more flexibility to GAT-QE such that it can learn whichever language structure is best fit to predict the answer from the question words, i.e., the task-specific syntax,
 2) Since all the nodes are first-order neighbors of each other, using a single layer GAT itself, the GAT-QE can encode the syntactic information from the entire question. This significantly reduces the number of parameters, and the question-encoder becomes light-weight.
3) Avoids any pre-processing to find the graph structure (depending on some off-the-shelf language parser) of the question upfront.  
Below, we explain GAT-QE in detail. 

\subsubsection{GAT-QE}
An illustration of the GAT-QE is shown in Figure~\ref{fig:gatqe}.
The input to GAT-QE is a question-graph $G=(V,E)$ with a complete digraph (fully-connected directed graph) structure, where the node-set $V$ represents the words, and edge set $E$ denotes various syntactic connections between the words. An edge $e_{i,j} \in E$ represents the task-specific syntactic dependency from node $i$ to $j$. Each node $i\in V$ is associated with a label $l_i \in \{1,2,..,n\}$ ($n$ is the sequence length) that denotes the position of the corresponding word in the sequence, and a feature vector $q_i$ contains the word-embedding of the corresponding word. The node-feature vectors $q_i$s are initialized using pre-trained Glove word-embeddings~\cite{glove}. Given $G$, the GAT-QE encodes it into the question-context vector $q$ in two steps as explained below:

\textbf{Node update: } 
In this step, GAT-QE updates each node feature vector $q_i$ from $G$ into $q^{'}_{i}$ by aggregating the information from the neighborhood nodes using a GAT. During this node-update process, the GAT assigns various task-specific syntactic weights to each edge $e_{i,j}$ in $G$. 
Specifically, the GAT finds a neighborhood attention vector $\alpha_i \in \mathcal{R}^{|\mathcal{N}_i|}$ for each node $i$ ($\mathcal{N}_i$ denotes the neighborhood of node $i$), where each value $\alpha_{i,j}\in \alpha_i$ represents the weight of the edge $e_{i,j}$, indicating the importance of the syntactic dependency from node $i$ to $j$. Mathematically, the $i^{th}$ node neighborhood attention vector $\alpha_i$  is defined as:

\begin{align}
    \alpha_i&= softmax(s_i) \nonumber \\
    s_{i,j}&= \text{LeakyReLU}(w_a^T[W_1^T q_i \Vert W_2^T q_j])
    \label{eq:att}
\end{align}

where $s_i \in \mathcal{R}^{|\mathcal{N}_i|}$ is the un-normalized neighborhood attention score vector for node $i$ and $s_{i,j}$ is the score for the dependency from node $i$ to $j$, $w_a \in \mathcal{R}^{2d_a}$ and $\{W_1,W_2\} \in \mathcal{R}^{d_{\omega} \times d_a}$ are learnable parameters and $\Vert$ denotes concatenation. The vectors $q_i$ and $q_j$ are the $i^{th}$ and $j^{th}$ node feature vectors respectively.

To stabilize the attention process, GAT uses multi-head attention inspired from Transformer~\cite{transformer} but in a different manner.
Specifically, the Transformer \cite{transformer} divides every input feature vector into $K$ (number of attention heads) chunks, and the ``scaled dot-product attention" (see Figure~\ref{fig:transformer}) is individually applied over each chunk.  In contrast, in GAT, every node feature is copied $K$ times, and then the attention mechanism is individually applied over each copy~\cite{gat}. To be more specific, in GAT-QE, $K$ attention heads (we use only 2 heads, i.e., $K=2$) run in parallel (using the same formulation as in Eq.~\eqref{eq:att}) resulting in a set of neighborhood attention weight vectors $\{\alpha_i^k; 1 \le k \le K \}$ for each node $i$. The GAT-QE uses these $K$ attention vectors to update the corresponding node feature vector $q_i$ to  $q_i^{'}\in \mathcal{R}^{d_q}$, as defined below:

\begin{align}
    q_i^{'}&= \frac{1}{K} \sum_{k=1}^{K} \sum_{j \in \mathcal{N}_i} \alpha_{i,j}^{k}* (W_g^k(q_j)) 
\end{align}
where $\mathcal{N}_i$ is the neighborhood of node $i$, $W_g^k$ is the learnable parameters of a transformation network that transforms the node-features from $d_{\omega}$ to $d_{q}$ dimensional spaces, $\alpha_{i,j}^k$ is the task-specific syntactic-dependency weight from node $i$ to $j$ in the $k^{th}$ attention head and $*$ denotes multiplication.

\textbf{Position-aware node aggregation: }
This step is where GAT-QE encodes the word-order information from the question.  First, all the updated node features ($q_i^{'}$) are rearranged in the same sequential order as in the input question using the node-labels ($l_i$) associated with each node. This process generates the sequence $Q^{'}=\{q_1^{'},q_2^{'},..,q_n^{'}\}$ where $n$ is the number of words in the question. Then we convolve $Q^{'}$ using 1-dimensional (1-D) temporal filters with window size $s$ to encode the word-order information. 
This 1-D CNN step results in the sequence $\hat{Q}=\{\hat{q}_1,\hat{q}_2,..,\hat{q}_n\}$, where $\hat{q}_i \in \hat{Q}$ is the convolved feature vector corresponds to the $i^{th}$ word which is defined as:
\begin{align}
    \hat{q_i}&=F_{conv1D}(q^{'}_{i:i+s-1},\theta) 
    \label{eq:conv1d}
\end{align}
where $F_{conv1D}$ is the 1-D CNN with window size $s$ (we use $s=3$) with $\theta$ as the learnable parameters. Note that the sequence $Q^{'}$ is appropriately $0$-padded before feeding into the 1-D CNN to maintain the length of the sequence after convolution.
The convolution step is followed by a sum-pooling operation over the sequence $\hat{Q}$ resulting the encoded question, i.e., $q=\sum_{i=1}^{n}\hat{q_i}$. 

\section{Experiments and Results}
\subsection{Experimental setup}
\textbf{Baseline: } We use the UpDn model by~\cite{bottomup} as the baseline for various experiments. We chose this model, since it is the widely-used baseline for OOD evaluation in VQA~\cite{overcoming,self_critical_bias,rubi,bias_aaai,hint,css}.

The UpDn model is based on question-guided object-level visual attention mechanism. The image is represented as a set of object-level CNN features, and the question-encoder is the GRU-QE as explained in Sec.~\ref{sec_qe}. The encoded question is used to find the relevant objects from the image via the visual attention mechanism.  The attended object-level CNN feature vector is considered as the encoded image vector. The encoded image and question vectors are then fused using element-wise multiplication to find a joint embedding vector.  This vector is then passed to the answer prediction network.

In the UpDn model, we supplant the existing question-encoder (i.e., GRU-QE) with the different types of encoders, as explained in Sec.~\ref{sec_qe}, for various experiments.  

\begin{table*}
\small
     \centering
     \caption{Out-of-Distribution (VQA-CPv2) and In-distribution (VQAv2-val) performance comparison among various question encoders on the UpDn baseline model.  QE denotes question-encoder.}
     \begin{tabular}{l|ccccccccc}
     \toprule
          &\multicolumn{4}{c}{Out-of-Distribution (VQA-CPv2)  }&&\multicolumn{4}{c}{In-distribution (VQAv2-val) }
          \\
          \cline{2-5}\cline{7-10}
          QE&Overall&Yes/No&Number&Other&&Overall&Yes/No &Number&Other\\
          \midrule
          GRU-QE&39.87& 44.95&12.97&44.22&&63.33&79.23 &43.15& 55.75\\
          BERT-QE&40.16&43.87&13.20&45.41&&65.42&82.29&43.47&57.12\\
          MMT-QE&41.05&43.46&13.54&46.22&&\textbf{65.75}&\textbf{82.67}&\textbf{44.46}&\textbf{57.56}\\
          
          GAT-QE&\textbf{45.88}&\textbf{59.03}&\textbf{18.34}&\textbf{46.43}&&62.51&77.48&44.41&55.93\\
          \bottomrule
     \end{tabular}
     \label{tab:qe_results}
 \end{table*}

\textbf{Datasets:} 
We use the VQA-CPv2~\cite{gvqa} dataset to evaluate the OOD performance of the models.  This dataset is constructed to test VQA models in an OOD setting, it contains different train and test set language-priors. Hence, a model that over-relies on the train set language-priors will show poor performance while testing. Currently, this dataset is widely-used to evaluate the OOD performance of VQA models~\cite{rubi,overcoming,self_critical_bias,vgqe,mutant}. 
For the in-distribution performance evaluation, we use the VQAv2 dataset~\cite{vqa2}. This dataset is one of the most popular datasets in VQA, and it contains similar language-priors among the train and test sets.  We use the standard VQA accuracy as the evaluation metric~\cite{vqa1} . 

\textbf{Implementation:} All of our implementations are using the PyTorch deep-learning framework~\cite{pytorch}. To allow easy reproducibility, we use the publicly available implementations for all our experiments. The details are described below. 

\textit{Image features:} 
We use the same object-level image features as used in the baseline model~\cite{bottomup}. These features are extracted from pre-trained Faster-RCNN with ResNet-101 backbone~\cite{bottomup}. 

\textit{Existing question-encoders:}
For the GRU-QE, BERT-QE, and MMT-QE, we use the publicly available implementations.  Specifically, we use the official GRU implementation from PyTorch~\cite{pytorch} for the GRU-QE, for the pre-trained ``BERT-base" (i.e., the BERT-QE) we use the implementation from~\cite{hugging_face} and for the MMT-QE (i.e., question-encoder of LXMERT), we use the code provided by the authors~\cite{lxmert}.  

\textit{GAT-QE:} We use the GloVe~\cite{glove} word-embeddings to initialize the question graph. We use the PyTorch-Geometric library~\cite{pytorch_geometry} to implement the  Graph attention network (GAT). We use the LeakyReLU with a negative slope of 0.2 as the non-linearity function. We use two heads to stabilize the neighborhood attention process, i.e., $K=2$. We fix the 1-D CNN window size as three, i.e., $s=3$. We fix the dimensions $d_q=d_a=512$.  

\textit{Training:}  We fix the answer vocabulary size as $3000$ following prior works~\cite{rubi,overcoming}. For all the experiments, we use similar training and optimization settings. We train the models using the Adamw~\cite{adamw} optimizer, with the Binary-cross entropy loss. We use a batch size of $128$. We use a fixed learning rate of $0.0002$ for GRU-QE and GAT-QE. In the case of BERT-QE and MMT-QE, we use a smaller learning rate of $0.00002$ as we follow the fine-tuning approach. 

\subsection{Performance analysis among question encoders}
In Table~\ref{tab:qe_results}, we show the performance of the UpDn model with various question-encoders on the OOD (VQA-CPv2) and in-distribution (VQAv2-val) test sets. We can see that the question-encoder has a significant role in the generalizability of the model (measured in terms of OOD performance).  We noticed that the RNN-based GRU-QE shows comparatively poor OOD performance than the self-attention-based models (BERT-QE, MMT-QE, and GAT-QE).  This is as expected because it has been shown that RNN-based question-encoders have an affinity towards the starting words of the question that often decides the question-type~\cite{analyzing_vqa, did_model}. Hence, the encoded question mainly contains the question-type information that encourages the model to learn spurious easy routes exploiting the language-priors (i.e., spurious correlations between the question-types and their most frequent answers in the train set). Upon comparing the OOD performance of self-attention-based models (i.e., BERT-QE, MMT-QE, and GAT-QE), the GAT-QE significantly outperformed its Transformer-based rivals (BERT-QE \& MMT-QE).  In the in-distribution case (i.e., on VQAv2-val), the Transformer-based MMT-QE outperformed all the others. To further analyze the question-encoders, we perform the following experiments.

 \subsection{Experiments on GRU-QE }
 In Table~\ref{tab:qe_results}, we observed that the GRU-QE shows comparatively poor OOD performance than the self-attention-based question-encoders.  Apart from the architectural pipeline, we note the following two differences in GRU-QE than the self-attention-based question-encoders (i.e., BERT-QE, MMT-QE, and GAT-QE). 1) GRU-QE uses the final cell hidden state vector as the encoded question, whereas the others use the sum-pooled word-level features, 2) GRU-QE is uni-directional, whereas the others are bi-directional question-encoders. Note that self-attention is a bi-directional operation as all the words get access to the past and future words in the sentence.  To analyze how the above two aspects affect the performance of GRU-QE, we perform the following experiments.
 
 \textbf{Sum-pooling aggregation: } In this experiment, as in the self-attention-based question-encoders, we apply the sum-pooling operation over the hidden states of the GRU cells in GRU-QE to get the question context vector $q$.  The results are shown in Table~\ref{tab:gru-qe-exp} (GRU-QE$_{sum}$).  We observed that sum-pooling aggregation improves both the OOD and in-distribution performances of GRU-QE.  
 
 \begin{table}
    \centering
    \small
    \caption{Performance comparisons of GRU-QE with sum-pooling aggregation and Bi-directional GRU (BiGRU-QE). The suffix \textit{sum} indicate the use of sum-pooling aggregation. }
    \begin{tabular}{lcc}
    \toprule
         QE & VQA-CPv2 & VQAv2-val \\
         \midrule
         GRU-QE & 39.87 & 63.33\\
         GRU-QE$_{sum}$& 41.02& \textbf{64.20} \\
         BiGRU-QE&40.30&62.76\\
         BiGRU-QE$_{sum}$&\textbf{41.60}&63.84\\
         \bottomrule
    \end{tabular}
    \label{tab:gru-qe-exp}
\end{table}

 \textbf{Bi-directional GRU-QE: } In this experiment, we use a bi-directional GRU in GRU-QE. A bi-directional GRU contains two uni-directional GRUs, one for processing the sequence from left to right and another for processing from right to left. Then, the final cell hidden state vectors of both the GRUs are concatenated to get the encoded question context vector $q$.  The results are shown in Table~\ref{tab:gru-qe-exp} (BiGRU-QE). With bi-directional GRU, we observed a slight increase in the OOD performance and a decrease in the in-distribution performance in GRU-QE.  
 
 \textbf{Bi-directional GRU-QE with sum-pooling: } This experiment combines the above two experiments. Specifically, we use a bi-directional GRU and then sum-pooled the word-level features from both the GRUs. Then we concatenated both the sum-pooled feature vectors to get the encoded question context vector $q$. The results are shown in Table~\ref{tab:gru-qe-exp} (BiGRU-QE$_{sum}$). Upon comparing with the GRU-QE, we can see a considerable increase in the OOD performance (i.e., from $39.87$ to $41.60$) and a slight increase in the in-distribution performance (i.e., from $63.33$ to $63.84$) with this setting.  
 
 The above experiments reveal that incorporating bi-directionality and sum-pooling aggregation to the RNN-based GRU-QE improves its OOD performance (i.e., 39.87 to 41.60).   However, upon comparing this performance improvement with the OOD performance of the self-attention-based GAT-QE, we can see that the GAT-QE significantly outperforms the bi-directional GRU-QE with sum-pooling aggregation (i.e., GAT-QE: 45.88 vs. BiGRU-QE$_{sum}$: 41.60, on VQA-CPv2).

\subsection{Experiments on self-attention based question-encoders}
  In Table~\ref{tab:qe_results}, we observed that among the self-attention-based question-encoders, the GAT-QE significantly outperformed the Transformer-based question-encoders (BERT-QE and MMT-QE) on the OOD case (i.e., on VQA-CPv2), whereas, in the in-distribution case (i.e., on VQAv2-val), the Transformer-based question-encoders performed better than the GAT-QE. Among the Transformer-based question-encoders, we noticed that the Vision-Language (VL) pre-trained MMT-QE performed better than the Language-only pre-trained BERT-QE.   To analyze these observations further, we perform the following experiments on the self-attention-based question-encoders.

   \begin{table}
 \small
     \centering
     \caption{Effect of ``position-encodings" (Pos-enc) on the OOD (VQA-CPv2) and in-distribution (VQAv2-val) performances of BERT-QE and MMT-QE. \ding{51} in the ``Pos-enc" denotes the presence of ``position-encodings" and \ding{55} vice-versa. ``+1D CNN" denote the use of 1D CNN instead of ``position-encodings". }
         \setlength\tabcolsep{2.5pt}
     \begin{tabular}{l|c|c|c}
          \toprule
          QE&Pos-enc&VQA-CPv2&VQAv2-val\\
          \midrule
          BERT-QE& \ding{51}&40.16& 65.42\\
          BERT-QE&\ding{55}&40.69& 65.00\\
          BERT-QE+1D CNN&\ding{55}&41.10& 65.29\\
          \midrule
          MMT-QE &\ding{51}&41.05& 65.75\\
          MMT-QE&\ding{55} &41.77& 64.91\\
          MMT-QE+1D CNN&\ding{55} &41.92& 65.22\\
          \midrule
            GAT-QE &\ding{55}& 45.88 &62.51\\
          \bottomrule
     \end{tabular}
     \label{tab:word_order}
 \end{table}
\textbf{Effect of position-encodings:}
Recall that both BERT-QE \& MMT-QE encode the word-order information using the ``position-encodings".  That is, learned vectors that are adding to the input token embeddings to indicate the absolute position of each token in the sequence. In this experiment, we evaluate the contribution of ``position-encodings" on the performance of BERT-QE and MMT-QE.  Concretely, we experiment by removing the ``position-encodings".

   The results are shown in Table~\ref{tab:word_order}.  We can see that, in both BERT-QE and MMT-QE, removal of ``position-encodings" increases the OOD performance (i.e., on VQA-CPv2). On the other hand, it decreases the in-distribution performance (i.e., on VQAv2-val).
   A possible reason for this behavior is that the question-encoders might have used the ``position-encodings" to give more weightage to the beginning-patterns (e.g., question-type patterns, i.e., what color is, how many, etc.) that encourages the model to learn spurious easy routes exploiting the language-priors.  Since the in-distribution test set contains similar language-priors as in the train set, the model can exploit these learned easy routes to get benefits on such test sets.

\textbf{Position-encodings vs. 1-D CNN:} 
Recall that in GAT-QE, we proposed to use 1-D CNN to encode word-order information. In this experiment, we evaluate the use of 1-D CNN instead of ``position-encodings" in BERT-QE and MMT-QE. Specifically, we removed the ``position-encodings" and used a 1-D CNN over the final-layer Transformer outputs. The results are shown in Table~\ref{tab:word_order}. We can see that supplanting ``position-encodings" with 1-D CNN improves the OOD performance of both BERT-QE (from 40.16 to 41.02) and MMT-QE (from 41.05 to 41.92).

To further analyze and make a quantitative comparison between ``position-encodings" and 1-D CNN in learning the language-priors, we conduct another experiment as described below.   Recall that language-priors in VQA significantly occur from the spurious correlations between the question-types and their most frequent answers in the train set~\cite{gvqa}. Hence, evaluating the trained models on in-distribution test sets that contain similar language-priors using only the question-type becomes a valid measure of the learned bias.  Thus, to evaluate how much the model benefited from the language-priors, we took a trained model and provided only the question-type information instead of the whole question, then evaluated the performance on the in-distribution test set (i.e., on VQAv2-val).  E.g., we provided only the ``what color is" pattern for the question ``what color is the car?".

 \begin{table}
 \small
     \centering
     \caption{In-distribution performance comparison of MMT-QE and BERT-QE using full question (Full Q) and only question-type (Q-type). ``Pos-enc" denotes the position encoding and \ding{51} represents its presence and \ding{55} vice versa. ``$\%$ Acc" measures the contribution of the Q-type to the full-Q accuracy (\textbf{Lower} is better). ``$\Delta$ Gap" denotes the absolute difference between the full-Q accuracy and the Q-type accuracy (\textbf{Higher} is better).
     }
     \setlength\tabcolsep{1.5pt}
     \begin{tabular}{l|c|c|c|c}
     \toprule
     QE&\makecell{Pos\\enc}&\makecell{VQAv2-val\\(overall)}&\makecell{\% Acc}&\makecell{$\Delta$ Gap}\\
     \midrule
          BERT-QE (Full Q)&\ding{51}&65.42&\multirow{2}{*}{60.16 }&\multirow{2}{*}{26.06 }\\
          BERT-QE (Q-type)&\ding{51}&39.36&\\
          \midrule
          BERT-QE (Full Q)&\ding{55}&65.00&\multirow{2}{*}{55.32}&\multirow{2}{*}{29.04 }\\
          BERT-QE (Q-type)&\ding{55}&35.96 &\\
          \midrule
          BERT-QE+1D CNN (Full Q)&\ding{55}&65.29&\multirow{2}{*}{57.52}&\multirow{2}{*}{27.77 }\\
          BERT-QE+1D CNN (Q-type)&\ding{55}&37.52 &\\
          
          \midrule
          \midrule
          MMT-QE (Full Q) & \ding{51}&65.75&\multirow{2}{*}{60.10}&\multirow{2}{*}{26.23}\\
          MMT-QE (Q-type)&\ding{51}&39.52&\\
          \midrule
          MMT-QE (Full Q)&\ding{55}&64.91&\multirow{2}{*}{52.68}&\multirow{2}{*}{30.71 }\\
          MMT-QE (Q-type)&\ding{55}& 34.20&\\
          \midrule
          MMT-QE+1D CNN (Full Q) & \ding{55}&65.22&\multirow{2}{*}{56.63}&\multirow{2}{*}{27.93}\\
          MMT-QE+1D CNN (Q-type)&\ding{55}&37.29&\\
         \bottomrule
     \end{tabular}
     \label{tab:qtype}
 \end{table}
 
The results are shown in Table~\ref{tab:qtype}. For a quantitative comparison, in Table~\ref{tab:qtype}, we define two metrics ``$\%$ Acc" and ``$\Delta$ Gap". The ``$\%$ Acc" denotes the percentage of contribution of the ``Q-type" accuracy to the ``Full-Q" accuracy.  A lower ``$\%$ Acc" indicates a lesser contribution from the Q-type and a higher value indicates vice-versa.  The ``$\Delta$ Gap" indicates the absolute difference between the ``Full-Q" and the ``Q-type" accuracy values.  A lower ``$\Delta$ Gap" indicates the over-reliance of the model on the Q-type and a higher value indicates vice-versa.

From the results in Table~\ref{tab:qtype}, we can see that upon using the ``position-encodings",  a major portion of the overall performance is coming from the question-type, i.e., using the spurious easy routes exploiting the language-priors, and it is comparatively less without using ``position-encodings". Also, we can see that using 1-D CNN instead of ``position-encodings" reduces the contribution from the language-priors.  For instance, consider the case of MMT-QE in Table~\ref{tab:qtype}; with ``position-encodings",  $60.10 \%$ of the overall accuracy is obtained using the question-type information only, and with 1-D CNN, it is reduced to $56.63 \%$.  Also, in MMT-QE, with 1-D CNN, the ``$\Delta$ Gap" is comparatively larger than using ``position-encodings" (i.e., 27.93 vs. 26.23).  
These results further clarify that ``position-encodings" encourage the model to learn the spurious easy routes exploiting language-priors, and 1-D CNN reduces this effect.  This indicates that, upon considering the generalizability of the model,  the 1-D CNN is a better alternative to ``position-encodings"  to encode the word-order information from the question.

\textbf{Effect of Vision-Language pre-training:}
In VQA, an emerging trend in various recent best-performing models (i.e., in terms of in-distribution performance) is the use of MMT-QE as the question-encoder, i.e., Vision-Language (VL) pre-trained Transformer-based question-encoders~\cite{lxmert,vilbert,vlbert}.  
In this experiment, we evaluate how much the pre-trained VL information contributes to the performance of the question-encoder in the OOD and in-distribution cases.  We experiment without using the VL pre-trained weights in MMT-QE. Specifically, we use MMT-QE initialized using BERT-base.  Note that this experiment is similar to a 9-layered version of the Language-only pre-trained BERT-QE. The results are shown in Table~\ref{tab:qe_size_pre-train}. We can see that without using the VL pre-trained weights, the MMT-QE significantly reduces the in-distribution performance (i.e., 65.75 to 64.72), whereas, on the other side, it shows only a slight decrease in the OOD performance (i.e., 41.05 to 40.87). 

We infer from the above experiment that in VQA, using VL pre-trained large multi-layered Transformer-based question-encoders  (i.e., MMT-QEs) significantly favors the in-distribution case and does not contribute much to the OOD performance.  
 \begin{table}
 \small
    \centering
    \caption{Effect of vision-language pre-training (VL pre-train) in MMT-QE. \ding{51} in the ``VL Pre-train" denotes pre-trained on Vision-Language data and \ding{55} vice-versa. \#L denotes the number of layers. Best values are in bold.}
    \begin{tabular}{l|c|c|c|c}
         \toprule
         QE&VL pre-train&\#L&VQA-CPv2&VQAv2-val\\
         \midrule
         MMT-QE&\ding{51}&9&41.05&65.75 \\
         MMT-QE&\ding{55}&9&40.87&64.72\\
         \midrule
         GAT-QE&\ding{55}&\textbf{1}&\textbf{45.88}&62.51\\
         \bottomrule
    \end{tabular}
    \label{tab:qe_size_pre-train}
\end{table}

 \begin{table}
 \small
    \centering
    \caption{Effect of the number of layers  (\#L) in the question-encoder. \#P denotes the total number of parameters in the question-encoder.}
    \setlength\tabcolsep{2.6pt}
    \begin{tabular}{l|c|c|c|c}
         \toprule
         QE&\#L&\#P&VQA-CPv2&VQAv2-val\\
         \midrule
         BERT-QE&12&110M&40.16&65.42\\
         BERT-QE&1&32.5M&42.03&64.26\\
         \midrule
         MMT-QE&9&89.20M&41.05&\textbf{65.75}\\
         MMT-QE&1&32.5M&41.93&64.53\\
         
         \midrule
         GAT-QE&\textbf{1}&\textbf{2.04M}&\textbf{45.88}&62.51\\
         \bottomrule
    \end{tabular}
    \label{tab:qe_size}
\end{table}
  \textbf{Effect of the number of layers in the question-encoder:}
In the above experiments, we noted that the relatively smaller (in terms of the number of parameters) single-layered GAT-QE outperformed the multi-layered BERT-QE (12 layers) and MMT-QE (9 layers) in the OOD performance. On the other hand, both of them outperformed GAT-QE on the in-distribution case. In this experiment, we analyze the contribution of the number of layers in the question-encoder to the performance. Specifically, we compare the performances of the single-layered versions of BERT-QE and MMT-QE with GAT-QE.

The results are shown in Table~\ref{tab:qe_size}. We observed that the single-layered versions of BERT-QE and MMT-QE reduces the over-fitting to the language-priors and improves the OOD performance than their multi-layered versions. On the other side, we observed a reduction in their in-distribution performance.  This is as expected since decreasing the number of parameters will decrease the chances of over-fitting to the train set biases, thus increasing the OOD performance and vice versa. Further, in Table~\ref{tab:qe_size}, we noticed that the single-layered Transformer-based question-encoders (BERT-QE and MMT-QE) show better in-distribution performance but poor OOD performance than the single-layered GAT-based question-encoder, i.e., the GAT-QE. A possible reason for this is that the Transformer architecture might have encouraged the model to learn biases than GAT.  To further clarify this, we experiment as follows.

\textbf{Transformer vs. GAT:}
At a higher level, we can see some conceptual similarities between GAT-QE and the Transformer-based question-encoders (BERT-QE and MMT-QE). Both GAT~\cite{gat} and Transformer~\cite{transformer} architectures are based on the multi-head self-attention but formulated in different ways (see Figure~\ref{fig:transformer} and Eq.~\ref{eq:att}). Despite the conceptual similarities, we found that the GAT-based question-encoder (i.e., GAT-QE) shows better bias-resistance (measured in terms of the OOD performance) than its Transformer-based rivals. To further analyze this and make a quantitative comparison of GAT and Transformer architectures in the question-encoder performance, we experiment as follows. We experiment with a naive Transformer-based question-encoder (i.e., without using the pre-trained weights as in BERT-QE and MMT-QE) and using similar parameter settings as in GAT-QE. Specifically, we use a single-layered Transformer with two attention heads, hidden-state dimensions as 512, and pre-trained Glove~\cite{glove} embeddings as the input token embeddings. 

\begin{table}
\small
    \centering
    \caption{Performance comparison of Transformer using similar settings as in GAT-QE.  Glove~\cite{glove} embeddings are used for the word-level features. ``Pos-enc" denotes the ``position-encodings" and \ding{51} represents its presence and \ding{55} vice versa. ``+1D CNN" denote the use of 1D CNN instead of ``position-encodings". }
    \setlength\tabcolsep{1.5pt}
    \begin{tabular}{l|c|c|c}
    \toprule
    QE&Pos-enc&VQA-CPv2&VQAv2-val\\
    \midrule
    Transformer &\ding{51}&40.82&64.34\\
    Transformer&\ding{55}&42.44&63.91\\
    Transformer + 1D CNN&\ding{55}&42.91&63.79\\
    \midrule
    GAT-QE &\ding{55}& 45.88 &62.51\\
    \bottomrule
    \end{tabular}
    \label{tab:gat_vs_transformer}
\end{table}

The results are shown in Table~\ref{tab:gat_vs_transformer}. We can see that the Transformer shows comparatively better in-distribution performance but poor OOD performance than GAT-QE. This indicates that the Transformer architecture encourages the model to learn biases from the train set than the GAT. Also,  in Table~\ref{tab:gat_vs_transformer}, we can see that, as observed in prior experiments on the Transformer-based question-encoders, the ``position encodings" significantly contribute to learning the bias, and 1-D CNN reduces this.

 To further analyze the above findings and probing the Transformer vs. GAT performance-gap reasons, we experiment as follows.  Going in-depth into these two architectures, we note the following two major differences between their implementations.  The Transformer and GAT use multi-head attention (MHA) in two different ways. Transformers divide the context vector into number-of-heads, and in contrast, GAT copies the context vector into number-of-heads times and performs attention. Another notable difference is in the scaled-dot-product attention (SDPA) in Transformer and Eq.~\eqref{eq:att} in GAT. The SDPA calculates the similarity scores using the dot-product between the context vectors, whereas Eq.~\eqref{eq:att} concatenates the context vectors and finds the scores by taking the dot-product with a learnable weight vector. We found the above two differences became advantageous to GAT-QE to show better generalizability. To demonstrate this, we change the above two aspects of GAT-QE with the Transformer counterparts. Results are shown in Table~\ref{tab:gat vs. Tr gap}. We can see that using the Transformer-like MHA and SDPA, the GAT-QE reduces the OOD performance but increases the in-distribution performance. These results provide further insights into the Transformer vs. GAT performance gap.
\begin{table}
    \centering
    \small
    \caption{Effect of GAT-like, Transformer-like (Tr-like) multi-head attention (MHA) and the scaled-dot-product attention (SDPA), on the performance of GAT-QE. }
    \begin{tabular}{lcc}
    \toprule
         Setup & VQA-CPv2 & VQAv2-val \\
         \midrule
         Tr-like MHA+ SDPA& 43.09 & 62.79 \\
         GAT-like MHA+SDPA & 43.84 & 62.84\\
         GAT-like MHA+ Eq.~\eqref{eq:att} & 45.88 & 62.51\\
         \bottomrule
    \end{tabular}
    \label{tab:gat vs. Tr gap}
\end{table}

\section{Conclusion}
This paper demonstrated that the sequence model in the question-encoder has a significant role in the generalizability of VQA models. We performed a detailed analysis of various existing RNN-based and Transformer-based question-encoders and proposed a novel GAT-based question-encoder, the GAT-QE, that shows comparatively better generalizability.    Our study analyzed various aspects of the question-encoder such as the architecture choice, number of parameters, Vision-Language pre-training, and methods to incorporate the word-order information.  This study reveals that a better sequence model in the question-encoder can improve the generalizability of VQA models even without using any additional relatively complex bias-mitigation techniques.  We hope this study could open various future research directions in developing robust VQA models resilient to language-priors.  A possible immediate benefit of this study could be the use of GAT-QE in applications where generalizability (in terms of OOD performance) and scalability (in terms of the memory consumption) of VQA models are a major concern. Apart from VQA, various findings from this study may also be beneficial to develop robust models in other domains that involve processing sequential data.
{\small
\bibliographystyle{ieee_fullname}
\bibliography{final}
}
\end{document}